\documentclass[runningheads]{llncs}
\usepackage{graphicx}
\usepackage{times}
\usepackage{latexsym}
\usepackage[T1]{fontenc}

\usepackage[utf8]{inputenc}
\usepackage{microtype}

\usepackage{indentfirst}
\usepackage{enumerate}
\usepackage{amsmath}
\usepackage{subcaption}
\usepackage{float}
\usepackage{hyperref}
\usepackage{helvet} 
\usepackage{courier}  
\usepackage{url}  
\usepackage{graphicx} 
\usepackage{wrapfig}
\usepackage[utf8]{inputenc}
\usepackage{booktabs}
\usepackage{amsmath}
\usepackage{amsfonts}
\usepackage{multirow}
\usepackage{color}
\usepackage[lined,boxed,commentsnumbered, ruled,linesnumbered]{algorithm2e}
\usepackage{microtype}
\usepackage{tabularx}
\usepackage{arydshln}
\usepackage{chngpage}
\usepackage{array}
\usepackage{hyperref}

\usepackage{microtype}
\usepackage{makecell}
\usepackage{tabularx}
\usepackage{booktabs}

\usepackage{marvosym}

\begin{document}

\title{HRoT: Hybrid prompt strategy and Retrieval of Thought for Table-Text Hybrid Question Answering}
%
%

\author{Tongxu Luo\inst{1,2} \and Fangyu Lei\inst{1,3} \and Jiahe Lei\inst{2} \and Weihao Liu\inst{1,2} \and   \\ 
Shizhu He\inst{1,3} \and Jun Zhao\inst{1,3} \and Kang Liu\inst{1,3}}
\institute{Institute of Automation, CAS \and
University of Science and Technology Beijing
\and University of Chinese Academy of Sciences}
%
\maketitle              
\begin{abstract}
Answering numerical questions over hybrid contents from the given tables and text~(TextTableQA) is a challenging task. Recently, Large Language Models (LLMs) have gained significant attention in the NLP community. With the emergence of large language models, In-Context Learning and Chain-of-Thought prompting have become two particularly popular research topics in this field. In this paper, we introduce a new prompting strategy called Hybrid prompt strategy and Retrieval of Thought for TextTableQA. Through In-Context Learning, we prompt the model to develop the ability of retrieval thinking when dealing with hybrid data. Our method achieves superior performance compared to the fully-supervised SOTA on the MultiHiertt dataset in the \textbf{few-shot} setting.


\keywords{ HybridQA \and Chain-of-Thought \and Language Models \and In-Context-Learning.}
\end{abstract}
\section{Introduction}
\label{introduction_section}
Question-answering (QA) systems aim to answer various questions using evidence located in structured knowledge bases, such as tables~\cite{pasupat2015compositional}\cite{yu2018spider} or unstructured texts~\cite{rajpurkar2016squad}. In real-world scenarios, QA systems often face the challenge of integrating various data resources of diverse types to answer complex questions, including numerical reasoning problems in financial statements. Therefore, the TextTableQA system, a hybrid of question answering over tables and texts ~\cite{chen2020open}\cite{chen2020hybridqa}\cite{chen2021finqa} has garnered increasing attention.

Recently, Large Language Models (LLMs) leverage Chain-of-Thought (CoT)~\cite{wei2022chain} prompts to break down complex problems into intermediate steps. Currently, there are three paradigms for mainstream CoT prompts. The first involves adding a single CoT trigger as a prompt for a single question, such as "Let's think step by step." This paradigm is called zero-shot, and in some simple datasets, LLMs perform well with this method. The second paradigm is manually constructing demonstrations, each consisting of a question, an inference process containing a CoT trigger, and a prompt to trigger the answer. The third paradigm is automatic demonstration selection and inference chain construction ~\cite{zhang2022automatic}. With zero-shot, LLMs generate inference chains for demonstrations one-by-one, then cluster and select typical demonstrations for few-shot.

\begin{figure}
\includegraphics[width=0.8\textwidth]{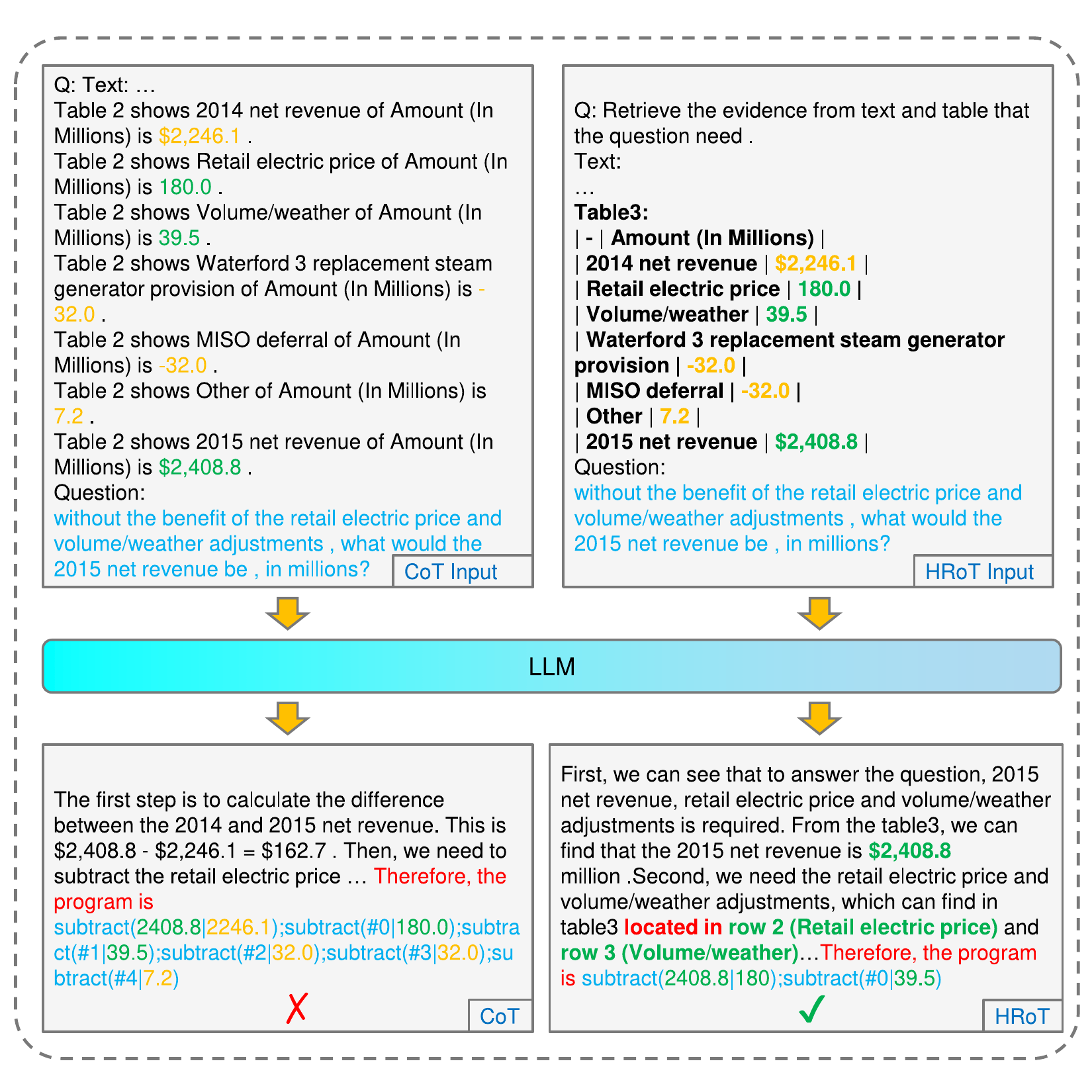}
\centering
\caption{Comparison result between CoT and HRoT with an example input. CoT uses text and table descriptions, which losing the \textbf{hierarchical information} of multi-hierarchical tables, while our method \textbf{reconstructs the tables} and constructs prompts to introduce \textbf{retrieval thinking} to prevent LLM from using irrelevant information for reasoning.}
\label{CoTandHRoT}
\end{figure}

We evaluated CoT on the MultiHiertt dataset~\cite{zhao2022multihiertt}, which contains long textual and multi-hierarchical tabular data in finance. The evaluation results showed that, while CoT has achieved State-Of-The-Art (SOTA) results on many datasets, it may not be effective for handling long and complex hybrid data that contains tables and text, especially when there is a lot of irrelevant information (as show in Fig \ref{CoTandHRoT}). The problem leads to poor performance of CoT is that CoT often relies on irrelevant information for reasoning, leading to incorrect reasoning chains and ultimately incorrect results. Furthermore, as shown in Fig~\ref{fig:table_eg}, tables in real scenes are typically hierarchical and complex, making it challenging to extract useful insights directly from the data. Therefore, it is necessary to address the problem of CoT being unable to retrieve correct evidence and explore effective modeling methods for real scene tables.

\begin{figure}[htbp]
\centering
\begin{subfigure}{0.45\textwidth}
\centering
\includegraphics[width=\textwidth]{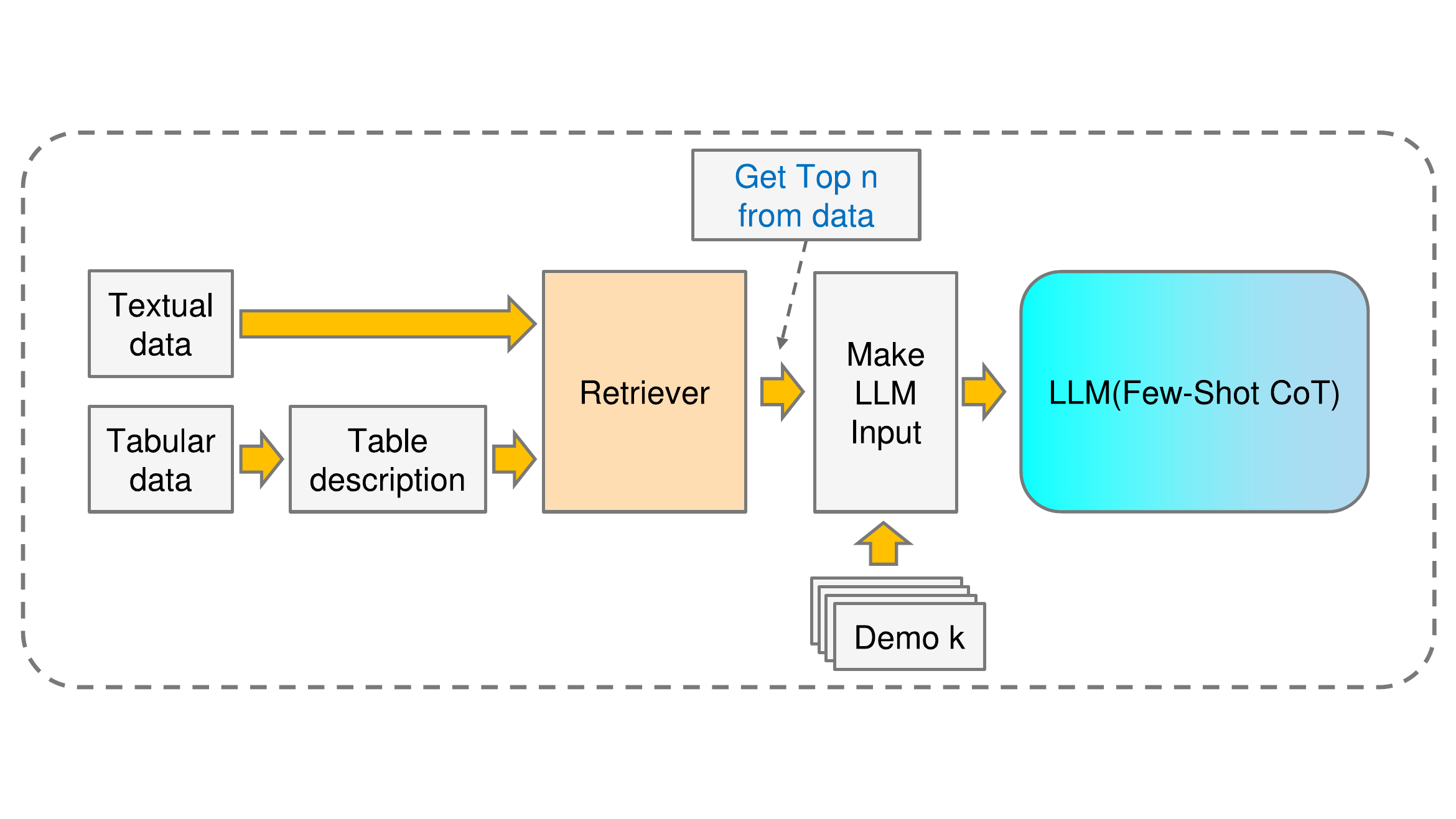}
\caption{The CoT pipline\label{CoT_pip}}
\end{subfigure}
\centering
\begin{subfigure}{0.45\textwidth}
\centering
\includegraphics[width=\textwidth]{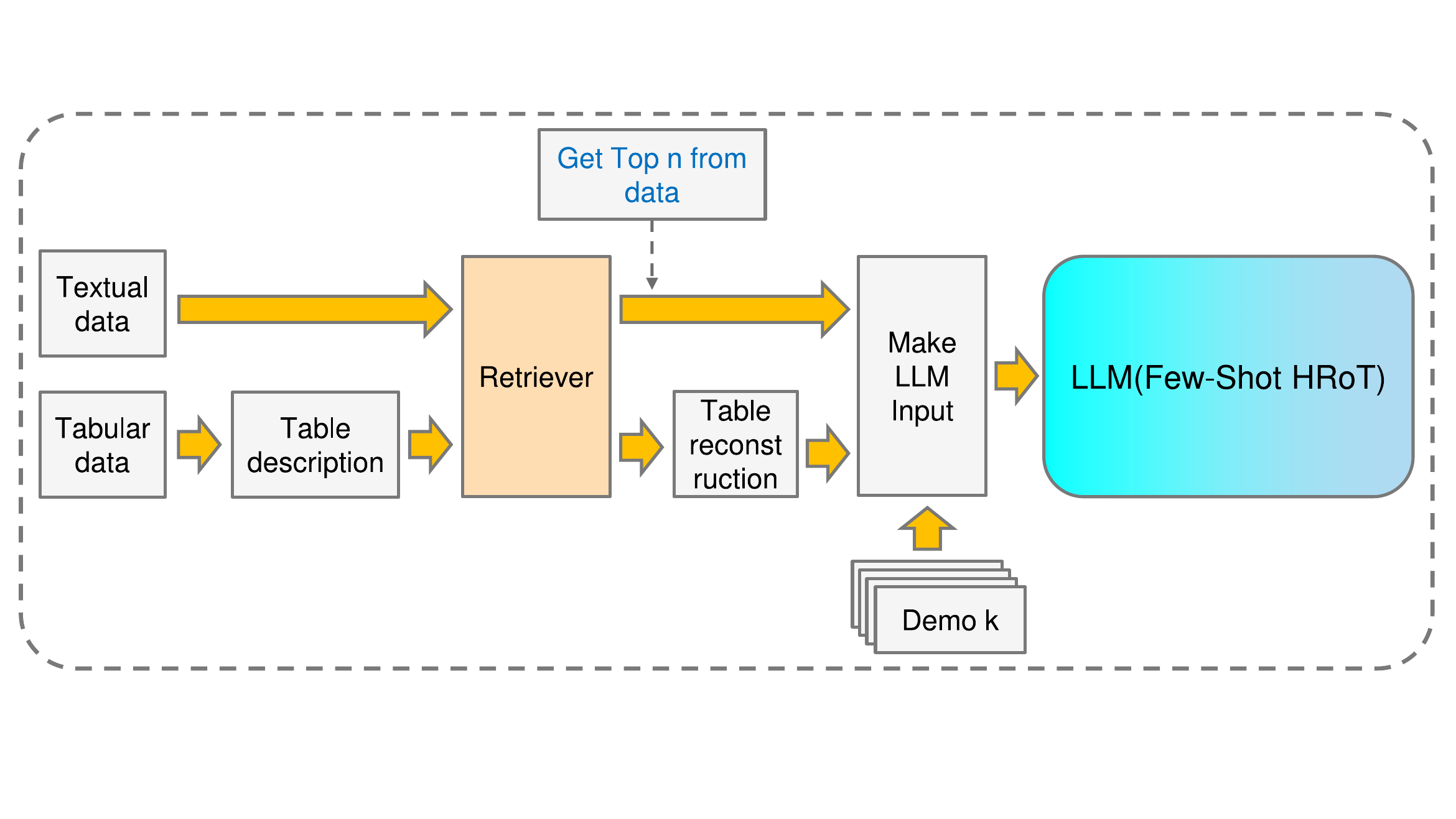}
\caption{The HRoT pipline\label{HRoT_pip}}
\end{subfigure}
\caption{The comparison between the pipline of CoT and HRoT in TextTableQA.}
\end{figure}

To address the aforementioned problems, we propose a novel method \textbf{HRoT}, which consists of two parts. Firstly, we introduce retrieval thinking by artificially constructing some arguments and guide the model to learn the way of thinking, which prevents LLM from relying on irrelevant information during reasoning. We illustrate the difference between CoT and our proposed method, \textbf{Retrieval of Thought}~(RoT), in Fig \ref{CoTandHRoT}. We provided several examples of retrieval-based thinking in the prompt. For example, ``We need to find ... located in ...''. Secondly, we propose a \textbf{Hybrid prompt strategy} that enhances the reasoning process by reconstructing the retrieved table based on the question type and considering the inherent hierarchical structure of the table, as described in Section~\ref{reconstruction_section}. The overall framework of the model is illustrated in Fig~\ref{HRoT_pip}. We compare CoT and HRoT in both zero-shot and few-shot settings, and our HRoT achieves better results in both settings, surpassing the fully supervised performance in 4-shots and achieving State-Of-The-Art performance\footnote{\url{https://codalab.lisn.upsaclay.fr/competitions/6738}}. 
In summary, our contributions are as follows:

(1) We propose a novel method to enhance the reasoning capability of models by introducing retrieval thinking. 

(2) Proposing Type-Aware Table Reconstruction algorithm to reconstruct multi-hierarchical tables based on the retrieved evidence.

(3) We propose a more powerful baseline retriever by introducing DeBERTa~\cite{hedeberta}.

\section{Related Work}

\subsection{In-Context Learning}
Large language models such as GPT-3 exhibit impressive few-shot learning ability~\cite{liu2023pre}\cite{dong2022survey}, requiring only a few questions and answers as prompts in the context without the need for finetuning on a dataset of training examples. However, this approach struggles with
tasks requiring complex reasoning~\cite{rae2021scaling}, leading researchers to explore prompting strategies. CoT~\cite{wei2022chain} is a chained reasoning approach that inserts a multi-step reasoning path before generating the final answer. Wang et al.~\cite{wang2022self} proposed a Self-Consistency decoding strategy to vote on the reasoning path, and Kojima et al.~\cite{kojima2022large} demonstrated that LLMs could act as zero-shot reasoners through the use of "Let’s think step-by-step". These methods focus on constructing inference chains, but cannot be well migrated to the field of HybridQA. CoT often relies on irrelevant information for reasoning, leading to incorrect reasoning chains and ultimately incorrect results. To overcome these challenges, our approach reconstructs a sub-table containing all evidence based on the retrieved table evidence, thus preserving the hierarchical information of multi-hierarchical tables. Moreover, we introduce retrieval thinking through prompts to prevent irrelevant information from being used in reasoning.
\subsection{TextTableQA}
In recent years, the hybrid form of question answering over tables and texts~(TextTableQA) has attracted more and more attention. There are two major question types for TextTableQA. The first is the fact reasoning question, whose answer is usually a span from the table or linked paragraphs, such as the contents in Wikipedia~\cite{chen2020open}\cite{chen2020hybridqa}. The second is the numerical reasoning question, which usually aims to use the contents of tables and texts for numerical calculation~\cite{zhu2021tat}\cite{chen2021finqa}. 

Our work focuses on numerical reasoning. TAT-QA~\cite{zhu2021tat}, FinQA~\cite{chen2021finqa} and Multihiertt~\cite{zhao2022multihiertt} are the numerical reasoning hybrid dataset which comes from the financial field. TAGOP~\cite{zhu2021tat} uses the sequence tagging method to extract facts, and performs a single arithmetic op- eration based on predefined operators. FinQANet~\cite{chen2021finqa} and MT2Net~\cite{zhao2022multihiertt} can perform multi-step reasoning, both of them use the LSTM decoder to autoregressively generate the program. UniRPG~\cite{zhou2022unirpg} generated numerical reasoning programs from tables to text, which can also use text spans as computation values. KIQA ~\cite{nararatwong-etal-2022-kiqa} through knowledge injection approach helpd the model to learn additional symbolic knowledge. RegHNT~\cite{lei2022answering}~\cite{wei2023multi} focused on designing a relation graph about the input. Different from the above methods, our method employs LLMs as reasoning module and introduces retrieval thinking through prompts, significantly improving the effectiveness of the retrieval process. S3HQA~\cite{lei2023s3hqa} and MMHQA~\cite{liu2023mmhqa} also use LLMs, but they focus on multi-hop TextTableQA tasks. 

\section{Method}
\subsection{Overview}
Our method is divided into three stages. The first stage is retrieval, which classifies the questions and retrieves for text and tables, selecting the top n as evidence. The second stage is reconstruct. In this stage, we first reconstruct the table of questions classified as arithmetic, and then use the text and reconstructed table as hybrid prompts to LLMs. The third stage is reasoning. We introduce retrieval thinking to guide the LLMs retrieve the evidence required for the reasoning from the text and table. The entire pipeline is shown in Figure \ref{HRoT_pip}
\subsection{Retriever}
Similar to the baseline, we use Pretrained Language Models (PLMs) to classify the questions into two types: arithmetic and span selection, and convert multi-hierarchical tables into table descriptions. However, unlike the baseline, we train separate models for text and table descriptions. In the training phase, for the $k$-th question $Q_k$, we have $N_k$ texts $P_k = \{p_1,p_2,\cdots ,p_{N_k}\}$ and $M_k$ table descriptions $T_k = \{t_1, t_2, \cdots, t_{M_k}\}$. We concatenate $Q_k$ with each $p_i$ and $t_i$(e.g., [CLS]\textbf{In what year is Home equity greater than 13000?}[SEP]\textbf{Annuities The following table presents the results of ...}[SEP]), and use DeBERTa~\cite{hedeberta} as the encoder to predict the correlation between the question and each text or table description pair. For DeBERTa's output:
\begin{equation} \label{deberta}
    H = [h_1;h_2;\cdots;h_l] = DeBERTa(X)
\end{equation}

Where $X$ is the concatenation of $Q_k$, $Q_k$ with $p_i$ or $Q_k$ with $t_i$. 

Among $H$, the classification information is $h_1$, and we use FFN as a classifier to binary classify the question's type or relevance.

During training, only a small portion of the given text contains the evidence required to answer a question. To address the problem of imbalanced positive and negative samples, we use resampling to increase the probability of sampling positive samples in a batch. Additionally, our loss function is defined as:
\begin{equation} \label{loss}
    Loss=CrossEntropy(y,\hat{y}) + \lambda \cdot DSCLoss(y,\hat{y})
\end{equation}

Where $\lambda$ is a hyperparameter and $DSCLoss$~\cite{li2019dice} is used to optimize the F1 score.

During the inference phase, we follow the same data processing steps as in the training phase. However, after predicting the relevance between the question and each text or table description pair, we sort $P_k$ and $T_k$ based on their relevance scores, and select the top $n$ and $m$ texts and table descriptions, respectively, as the retrieved candidate evidence.

\subsection{Hybrid Prompt Strategy}
\label{reconstruction_section}

Performing arithmetic operations on multi-hierarchical tables can present challenges when certain spatial information is not explicitly included in table descriptions. To address this issue, we propose a hybrid prompt strategy that utilizes hybrid data to prompt large language models (LLMs). Specifically, we introduce a type-aware table reconstruction algorithm to reconstructe large and complex tables to sub-tables.

\begin{figure}
\includegraphics[width=0.8\textwidth]{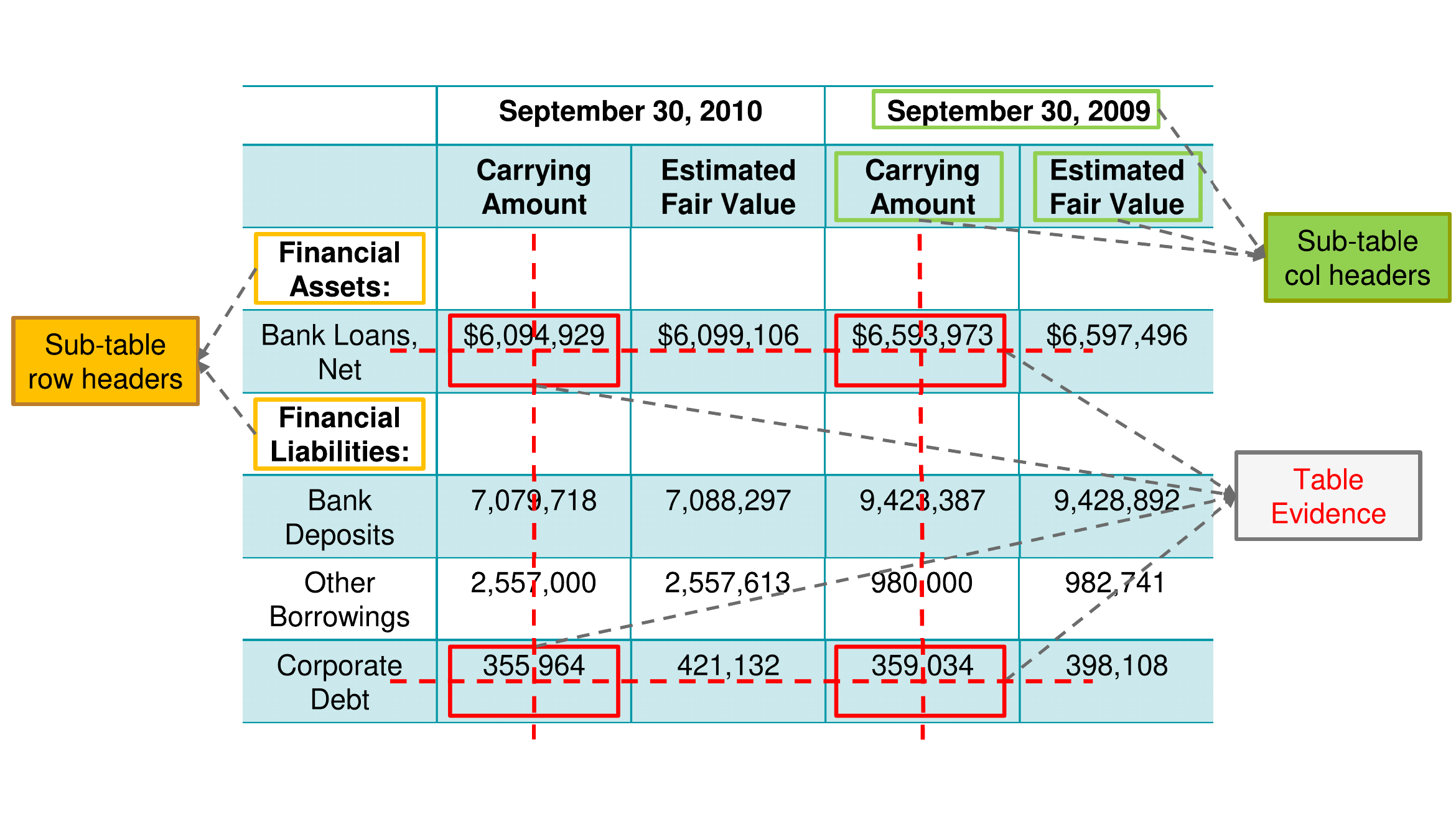}
\centering
\caption{An expamle of Multi-hierarchical table.}
\label{fig:table_eg}
\end{figure}

For example, consider the  MultiHiertt dataset in Fig \ref{fig:table_eg}, where each table contains hierarchical column and row headers. Ignoring the hierarchical structure of the headers may result in incorrect reasoning outcomes.

We first classify the questions into two types: arithmetic and span selection. Then, for one of arithmetic questions $q$, we get the tables of $q$ which is $T_q$. For one of table $t$ in $T_q$, we partition the table and obtain a span list $L$. For example, for the table in Fig \ref{fig:table_eg}, we obtain $L=[[0,1][2,3][4,7]]$, where $[0,1]$ represents the row span of the table header, and $[2,3]$ and $[4,7]$ represent the spans of sub-tables. Then, for each piece of evidence, we determine which row span it belongs to and retain the sub-header of that span. Therefore, the set $R$ of rows to be retained actually consists of three parts, $R=\{h_r, h_{sub}, r_e\}$. The set $C$ of columns to be retained consists of two parts, $C=\{h_c, c_e\}$. With the rows and columns to be retained identified, we can easily reconstruct the table.

Our table reconstruction algorithm can be summarized as follows:


\begin{algorithm}[H]
  \SetAlgoLined
  \KwIn{Question set $Q$;Table set $\{T_1, \cdots, T_n\}$;Evidence $\{E_1, \cdots, E_n\}$;}
  \For{$q$ in $Q$}{
  CLS = $Classifier(q)$\;
  \If{CLS is arithmetic}{
  \For{$t$ in $T_q$}{
    Get Table Span List $L_t$\;
    Use $L_t$, get headers $h_r$ and $h_c$ from $t$;\\
    \textbf{Initialize:}Row to reserve $R=\{h_r\}$;Col to reserve $C=\{h_c\}$;\\
    Get evidence $E_t$\;
    \For {$e$ in $E_t$}{
    Use $L_t$, get the Sub header $r$ and $c$ of $e$ from t\;
    Insert $r$ to $R$\;
    Insert $c$ to $C$\;
    }
    Update $t$ reserve the rows in $R$ and the cols in $C$\;
    }
    }
  }
  \KwOut{Reconstructed tables}
  \caption{Table Reconstruction}
  \label{alg:algorithm1}
\end{algorithm}

\subsection{Hybrid Retrieval of Thought}
To address the problem of LLM selecting incorrect evidence for reasoning, we introduce HRoT, which adopts a retrieval-based approach to gradually retrieve the evidence required for the question and generate the answer based on these evidence.


\subsubsection{HRoT Prompting for Zero-Shot}
\begin{figure}[htbp]
\centering
\begin{subfigure}{0.49\textwidth}
\centering
\includegraphics[width=\textwidth]{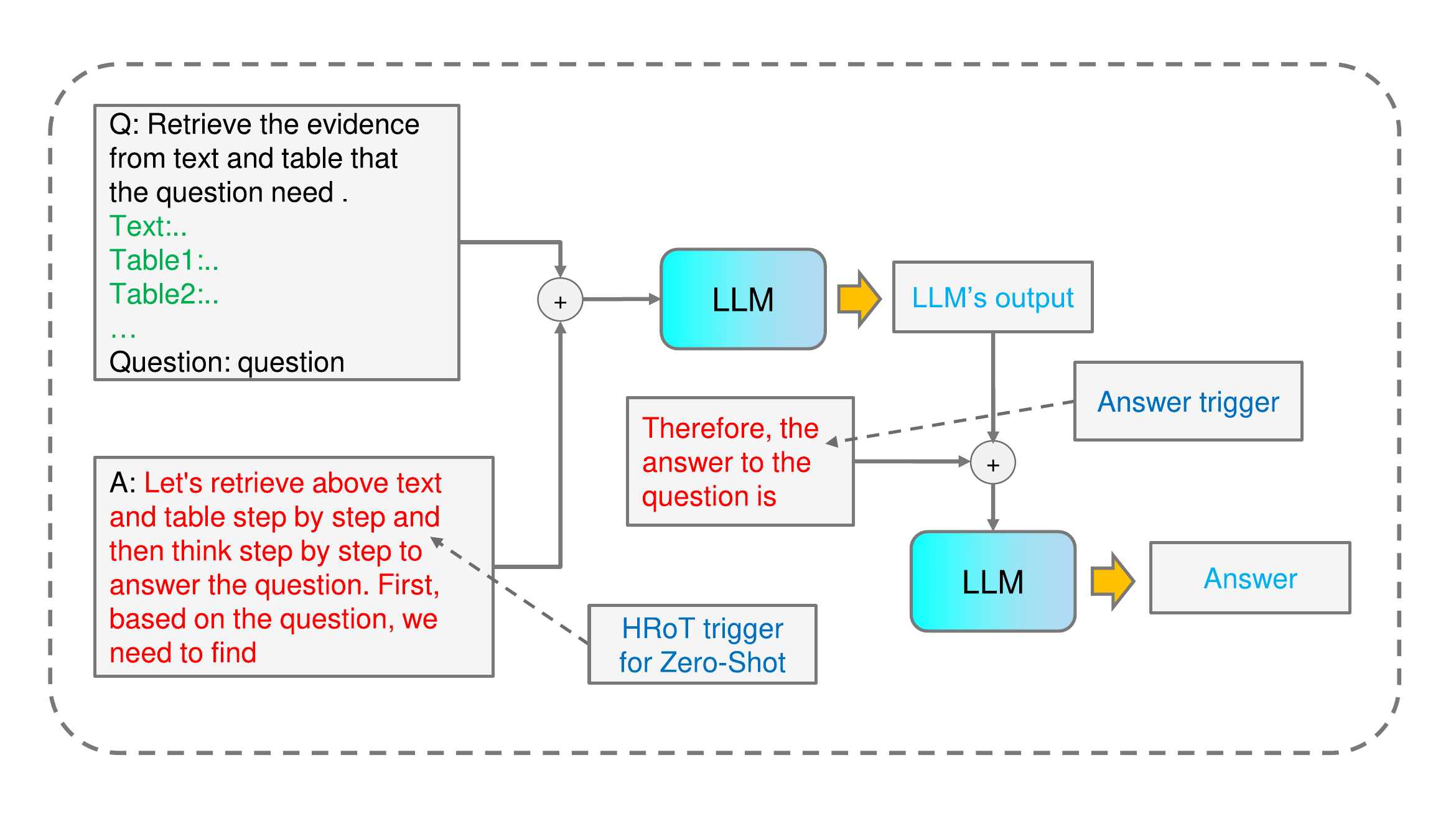}
\caption{For Zero-Shot Reasoning\label{zero_shot_hrot}}
\end{subfigure}
\centering
\begin{subfigure}{0.49\textwidth}
\centering
\includegraphics[width=\textwidth]{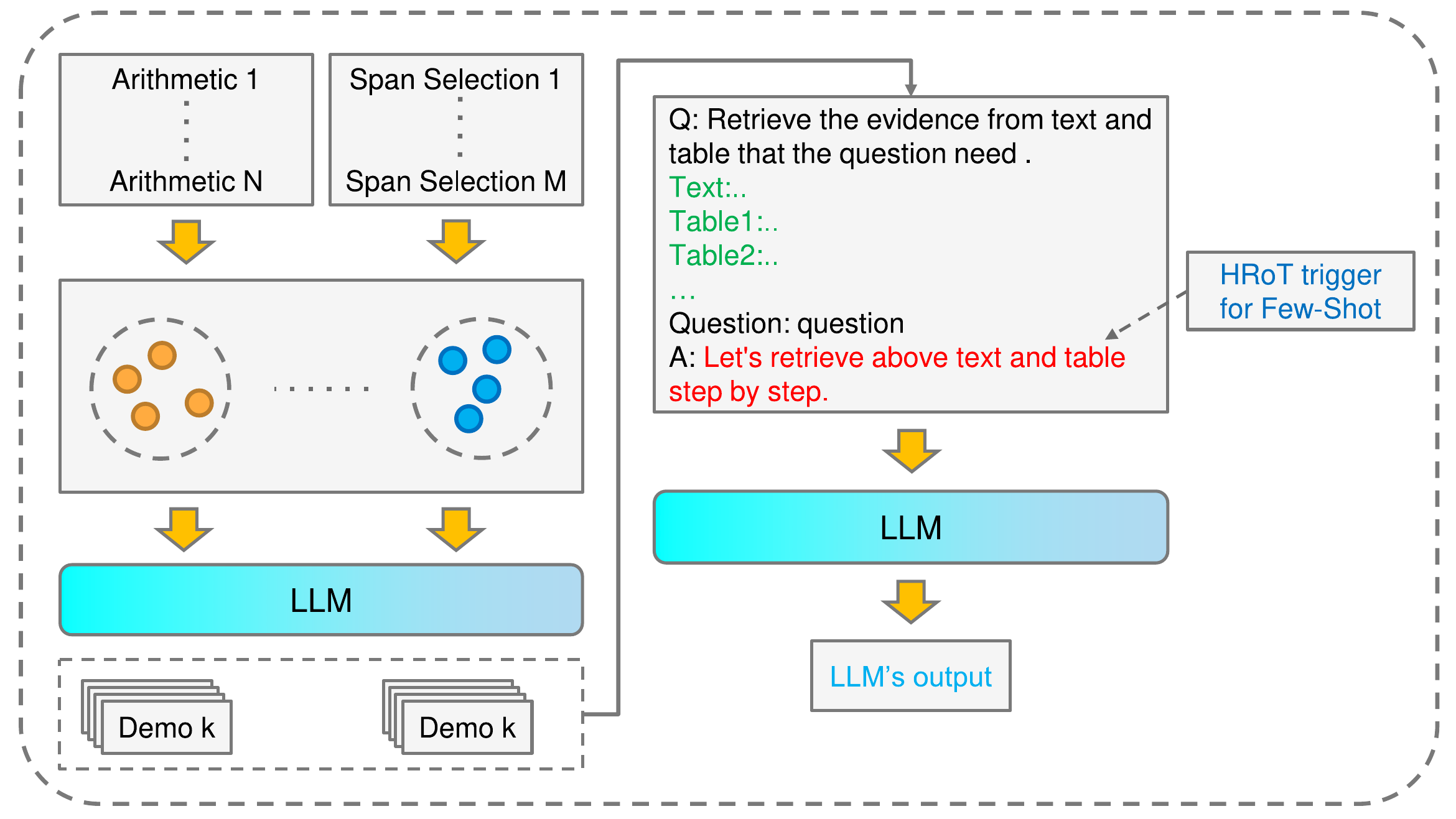}
\caption{For Few-Shot Reasoning\label{few_shot_hrot}}
\end{subfigure}
\caption{An example of HRoT. The Text in green is the retrieval result from Retriever and the Tables in green is the reconstructed tables.}
\end{figure}

For Zero-Shot reasoning, we use the prompt "Let's retrieve above text and table step by step and then think step by step to answer the question. First, based on the question, we need to find" to guide the LLM to conduct retrieval before answering the question. Finally, we use "Therefore, the answer to the question is" as the Answer trigger to extract the answer. The specific process is shown in Fig \ref{zero_shot_hrot}.


\subsubsection{HRoT Prompting for Few-Shot}
For Few-Shot, we adopted a similar approach to Auto-CoT~\cite{zhang2022automatic} and used clustering to select representative examples for demonstration, with a difference being that we clustered two types of questions separately, namely arithmetic and span selection. We first clustered the training set, then applied Zero-Shot on these examples. However, the correctness of the retrieved chains cannot be guaranteed, so we manually corrected any errors in the output. For Few-Shot, we directly used "Let's retrieve above text and table step by step" as the prompt. For arithmetic questions, we followed the requirements of the dataset and used the "Program" format as the answer. The specific process is shown in Fig \ref{few_shot_hrot}.


\section{Experiments}
\subsection{Datasets}
We conducted our experiments on the MultiHiertt dataset~\cite{zhao2022multihiertt}. Compared with existing datasets, each document in MultiHiertt contains multiple hierarchical tables and longer un- structured text. A more complex reasoning process across multiple tables and paragraphs is required to correctly answer the question (Zhao et al., 2022). The dataset consists of 10,440 questions with 2,513 financial documents, and is split into three parts: training (75\%), development (10\%), and test (15\%).
\subsection{Implementation Details}
During the retrieval phase, we tested several pre-trained language models (PLMs) including BERT, RoBERTa, and DeBERTa. For each PLM, we trained separate models for text and table descriptions. To address the problem of imbalanced positive and negative samples, we utilized resampling to increase the probability of sampling positive samples in a batch. Furthermore, to directly optimizes the F1 score, $\lambda$ in \ref{loss} is set to $0.5$.


During the reasoning phase, we used the OpenAI GPT3.5 (text-davanci-003) API with the setting $temperature=0$. We conducted experiments on CoT, HRoT without reconstructed table, and HRoT with reconstructed table under 0-4 shot settings.

\subsection{Main Results}


Table \ref{main_results} presents a comparison between our proposed method and several typical methods on the test set. As can be seen, our method significantly outperforms the previous baselines in terms of both EM and F1 scores. These results demonstrate the effectiveness of our approach in addressing complex, hierarchical table-based question answering tasks.


\begin{wraptable}{r}{0.35\textwidth}
  \vspace{-1.1cm}
  \caption{\label{main_results} Main results on test set.}
  \vspace{0.3cm}
  \setlength{\tabcolsep}{0.6mm}{
            \centering
         \begin{tabular}{ccc}
        \hline
                     & EM    & F1    \\ \hline
        TAGOP~\cite{chen2021finqa}        & 17.81 & 19.35 \\ \hline
        FinQANet~\cite{chen2021finqa}     & 31.72 & 33.60 \\ \hline
        MT2Net~\cite{zhao2022multihiertt}       & 36.22 & 38.43 \\ \hline
        NAPG~\cite{zhang2022napg}         & 44.19 & 44.81 \\ \hline
        HRoT-fewshot & \textbf{46.17} & \textbf{46.91} \\ \hline
        \end{tabular} } 
  \vspace{-0.6cm}
 \end{wraptable}

\subsection{Ablation Study}
We conducted ablation experiments on the development set to validate the effectiveness of retriever and on the test set to validate the effectiveness of Table Reconstruction and HRoT Prompting. Additionally, we compared the performance of HRoT under 0-4 shot settings. Table Reconstruction involves reconstructing a sub-table based on the retrieved results. HRoT Prompting is used to guide the LLM to retrieve the correct evidence from both text and table.

\begin{table}
	\caption{Ablation study on Bert, RoBERTa, DeBERTa and DeBERTa(+DSCLoss) Retriever using top 5 on text and top 10 on table setting.}\label{tab2}
	\centering
\begin{tabular}{lll}
\hline
             & Text Recall    & Table Recall    \\ \hline
Bert & 84.14 & 71.47 \\ \hline
RoBERTa         & 91.98 & 86.11 \\ \hline
DeBERTa       & 93.62 & 90.46 \\ \hline
DeBERTa(+DSCLoss)       & \textbf{94.48} & \textbf{91.27} \\ \hline
\end{tabular}
\end{table}

\subsubsection{Effect of proposed retriever}
As shown in Table \ref{tab2}, we validate the BERT, RoBERTa, DeBERTa and DeBERTa(+DSCLoss) Retriever. Comparing DeBERTa(+DSCLoss) with Bert, DeBERTa achieves an improvement of 10.34\% in the Text Recall, and 19.8\% in the Table Recall. When DSCLoss is removed, the Text Recall decreases by approximately 0.86\% and the Table Recall decreases by approximately 0.81\%.


\begin{table}
	\caption{Ablation study on Table Reconstruction and HRoT Prompting.}\label{tab3}
	\centering
\begin{tabular}{lll}
\hline
             & EM    & F1    \\ \hline
HRoT-fewshot & \textbf{46.17} & \textbf{46.91} \\ \hline
w/o Deberta~(w Roberta)   & 45.89 & 46.57  \\ \hline
w/o Hybrid Prompt Strategy         & 44.35 & 44.96 \\ \hline
CoT-fewshot       & 40.04 & 40.77 \\ \hline
\end{tabular}
\end{table}

\subsubsection{Effect of HRoT}
As shown in Table \ref{tab3}, when the reconstructed table is replaced with table descriptions under the same settings, the EM score decreases by approximately 1.82\% and the F1 score decreases by approximately 1.95\%. Comparing HRoT with CoT under the same settings, HRoT with reconstructed table achieves an improvement of 6.13\% in EM and 6.14\% in F1, while HRoT without reconstructed table achieves an improvement of 4.31\% in EM and 4.13\% in F1. These results demonstrate the effectiveness of our proposed improvements.

\begin{table}
	\caption{Experiments on different numbers of examples for HRoT and CoT.}\label{tab4}
	\centering
\begin{tabular}{l|cc|cc}
\hline
       & \multicolumn{2}{c|}{\textbf{HRoT}}                                         & \multicolumn{2}{c}{\textbf{CoT}} \\
       & \textbf{EM}                     & \textbf{F1}                     & \textbf{EM} & \textbf{F1} \\ \hline
0-shot & 22.67                           & 23.72                           & 20.45      & 21.67      \\
1-shot & 39.55                           & 40.62                           & 29.14      & 30.15      \\
2-shot & 41.53                           & 42.67                           & 34.76      & 35.49      \\
3-shot & 43.38                           & 44.43                           & 36.03      & 36.96      \\
4-shot &   \textbf{46.17} & \textbf{46.91} & 39.03      & 39.77      \\ \hline
\end{tabular}
\end{table}

\subsubsection{Different shots on HRoT}
As shown in Table \ref{tab4}, we used a 0-4 shot setting, and it can be seen that when doing few-shot, both EM and F1 scores are positively correlated with the number of demonstrations. When doing zero-shot, due to the inability of LLM to generate the required "Program" format for the dataset, there are more computational errors.

\section{Conclusion}
In this study, we investigate the construction of appropriate demonstrations and prompts for hybrid data of text and tables and propose HRoT, short for  Hybrid prompt strategy and Retrieval of Thought for TextTableQA. By significantly reducing the retrieval errors of evidence in hybrid data for LLMs, our method achieves SOTA performance on the MultiHiertt dataset.

%
%
%
\bibliographystyle{splncs04}

\bibliography{myref.bib}

\end{document}